\documentclass[runningheads]{llncs}
\usepackage{graphicx}
\usepackage{amsmath,amssymb} 
\usepackage{color}
\usepackage{url}
\usepackage{multirow}
\usepackage{tabularx}
\usepackage{tabulary}
\usepackage{booktabs}
\usepackage{amsfonts}
\usepackage{amsmath}
\usepackage{nicefrac}
\usepackage{microtype}
\usepackage{enumitem}
\usepackage{xcolor}
\usepackage{kotex}
\usepackage{algorithm, algpseudocode}%
\usepackage{subcaption}
\usepackage{array}
\usepackage{hyperref}
\usepackage{nicefrac}       
\usepackage{microtype}      
\usepackage{adjustbox}
\usepackage{multirow}
\usepackage{threeparttable}
\usepackage{makecell}
\usepackage{tabulary}
\usepackage{amsmath}
\usepackage{tabularx}
\usepackage{pifont}
\usepackage{kotex}
\usepackage{bbold}
\usepackage{wrapfig}

\usepackage{enumitem}
\usepackage{graphicx}
\usepackage{scalerel}
\usepackage{caption}
\usepackage{lipsum}

\begin{document}
\pagestyle{headings}
\mainmatter

\title{Patch SVDD: Patch-level SVDD \\ for Anomaly Detection and Segmentation}
\titlerunning{Patch SVDD: Patch-level SVDD for Anomaly Detection and Segmentation}
\author{Jihun Yi \and Sungroh Yoon}
\authorrunning{Jihun Yi}
\institute{
Department of Electrical and Computer Engineering, \\Seoul National University, Seoul, South Korea\\
\email{
\{
  \href{mailto:t080205@snu.ac.kr}{t080205},
  \href{mailto:sryoon@snu.ac.kr}{sryoon}
  \}@snu.ac.kr
}
}

\maketitle

\newcommand{\xvec}{\mathbf{x}}
\newcommand{\Xvec}{\mathbf{X}}
\newcommand{\pvec}{\mathbf{p}}
\newcommand{\Atheta}{\mathcal{A}_{\theta}}
\newcommand{\mmap}{\mathcal{M}}
\newcommand{\Loss}{\mathcal{L}}

\newcommand{\fbig}{f_{\text{big}}}
\newcommand{\fsmall}{f_{\text{small}}}

\newcommand{\textblue}{\textcolor{blue}}
\newcommand{\textred}{\textcolor{red}}
\newcommand{\xmark}{\text{\ding{55}}}
\newcommand{\cmark}{\text{\ding{51}}}

\newcommand{\etal}{\emph{et al.}}

\begin{abstract}
In this paper, we address the problem of image anomaly detection and segmentation.
Anomaly detection involves making a binary decision as to whether an input image contains an anomaly, and anomaly segmentation aims to locate the anomaly on the pixel level.
Support vector data description (SVDD) is a long-standing algorithm used for an anomaly detection, and we extend its deep learning variant to the patch-based method using self-supervised learning.
This extension enables anomaly segmentation and improves detection performance.
As a result, anomaly detection and segmentation performances measured in AUROC on MVTec AD dataset increased by 9.8\% and 7.0\%, respectively, compared to the previous state-of-the-art methods.
Our results indicate the efficacy of the proposed method and its potential for industrial application.
Detailed analysis of the proposed method offers insights regarding its behavior, and the code is available online\footnote{\url{https://github.com/nuclearboy95/Anomaly-Detection-PatchSVDD-PyTorch}}.
\end{abstract}
\section{Introduction}
Anomaly detection is a binary classification problem to determine whether an input contains an anomaly.
Detecting anomalies is a critical and long-standing problem faced by the manufacturing and financial industries.
Anomaly detection is usually formulated as a one-class classification because abnormal examples are either inaccessible or insufficient to model distribution during the training.
When concentrating on image data, detected anomalies can also be localized, and anomaly segmentation problem is to localize the anomalies at the pixel level.
In this study, we tackle the problems of image anomaly detection and segmentation.

One-class support vector machine (OC-SVM)~\cite{ocsvm} and support vector data description (SVDD)~\cite{svdd} are classic algorithms used for one-class classification.
Given a kernel function, OC-SVM seeks a max-margin hyperplane from the origin in the kernel space.
Likewise, SVDD searches for a data-enclosing hypersphere in the kernel space.
These approaches are closely related, and Vert et al.~\cite{ocsvm_consistency} showed their equivalence in the case of a Gaussian kernel.
Ruff et al.~\cite{deepSVDD} proposed a deep learning variant of SVDD, Deep SVDD, by deploying a deep neural network in the place of the kernel function.
The neural network was trained to extract a data-dependent representation, removing the need to choose an appropriate kernel function by hand.
Furthermore, Ruff et al.~\cite{deep_sad} re-interpreted Deep SVDD in an information-theoretic perspective and applied to semi-supervised scenarios.

\begin{figure}[t]
    \centering
    \includegraphics[width=\linewidth]{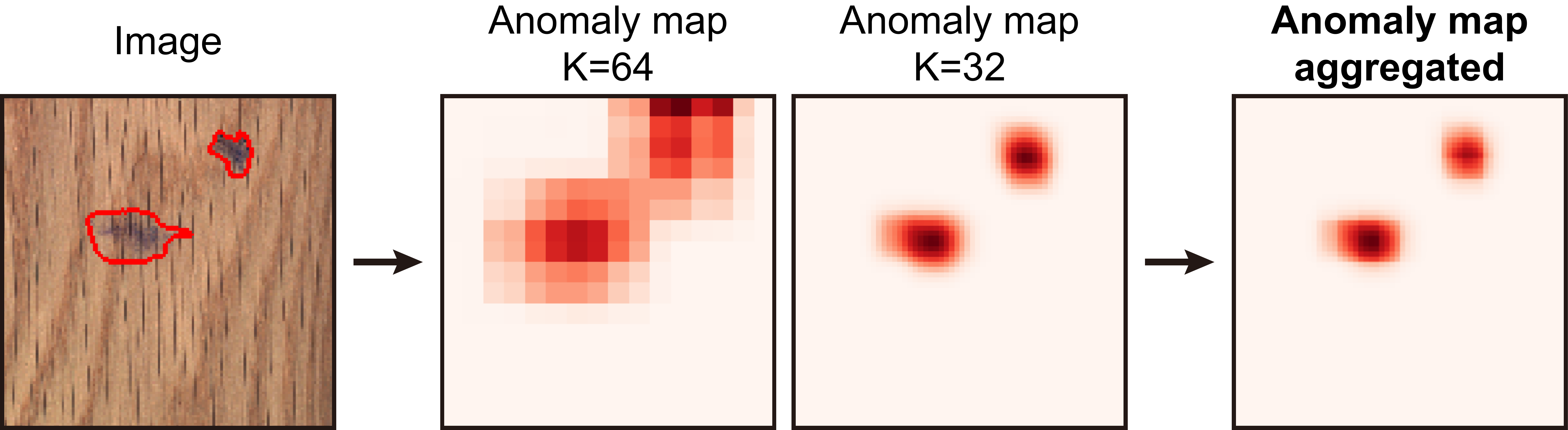}
    \vspace{-1.6em}
    \caption{
    \textbf{Proposed method localizes defects in an MVTec AD~\cite{mvtecad} image.}
    Patch SVDD performs multi-scale inspection and combines the results.
    As a result, the anomaly map pinpoints the defects (contoured with a red line).
    }
    \label{fig:example}
    \vspace{-1.4em}
\end{figure}

In this paper, we extend Deep SVDD to a patch-wise detection method, thereby proposing Patch SVDD.
This extension is rendered nontrivial by the relatively high level of intra-class variation of the patches and is facilitated by self-supervised learning.
The proposed method enables anomaly segmentation and improves anomaly detection performance.
Fig.~\ref{fig:example} shows an example of the localized anomalies using the proposed method.
In addition, the results in previous studies~\cite{patch_location,lens} show that the features of a randomly initialized encoder might be used to distinguish anomalies.
We detail the more in-depth behavior of random encoders and investigate the source of separability in the random features.

\section{Background}
\subsection{Anomaly detection and segmentation}
\subsubsection{Problem formulation}  
Anomaly detection is a problem to make a binary decision whether an input is an anomaly or not.
The definition of \textit{anomaly} ranges from a tiny defect to an out-of-distribution image.
We focus here on detecting a defect in an image.
A typical detection method involves training a scoring function, $\Atheta$, which measures the abnormality of an input.
At test time, inputs with high $\Atheta(\xvec)$ values are deemed to be an anomaly.
A \textit{de facto} standard metric for the scoring function is the area under the receiver operating characteristic curve (AUROC), as expressed in Eq.~\ref{eq:auroc}~\cite{auroc}.

\begin{equation} \label{eq:auroc}
    \text{AUROC}\left [ \Atheta \right ] = \mathbb{P} \left [ \Atheta(\Xvec_{\text{normal}}) < \Atheta(\Xvec_{\text{abnormal}}) \right ].
\end{equation}
A good scoring function is, thus, one that assigns a low anomaly score to normal data and a high anomaly score to abnormal data.
Anomaly segmentation problem is similarly formulated, with the generation of an anomaly score for every pixel (i.e., an anomaly map) and the measurement of AUROC with the pixels.

\subsubsection{Auto encoder-based methods}
Early deep learning approaches to anomaly detection used auto encoders~\cite{vae_ad,recon_and_detect,ocgan}.
These auto encoders were trained with the normal training data and did not provide accurate reconstruction of abnormal images.
Therefore, the difference between the reconstruction and the input indicated abnormality.
Further variants have been proposed to utilize structural similarity indices~\cite{ssim_ae}, adversarial training~\cite{recon_and_detect}, negative mining~\cite{ocgan}, and iterative projection~\cite{iterative_project}.
Certain previous works utilized the learned latent feature of the auto encoder for anomaly detection.
Akcay et al.~\cite{ganomaly} defined the reconstruction loss of the latent feature as an anomaly score, and Yarlagadda et al.~\cite{satellite} trained OC-SVM~\cite{ocsvm} using the latent features.
More recently, several methods have made use of factors other than reconstruction loss, such as restoration loss~\cite{itae} and an attention map~\cite{ve_vae}.

\begin{figure}[t]
    \centering
    \includegraphics[width=\textwidth]{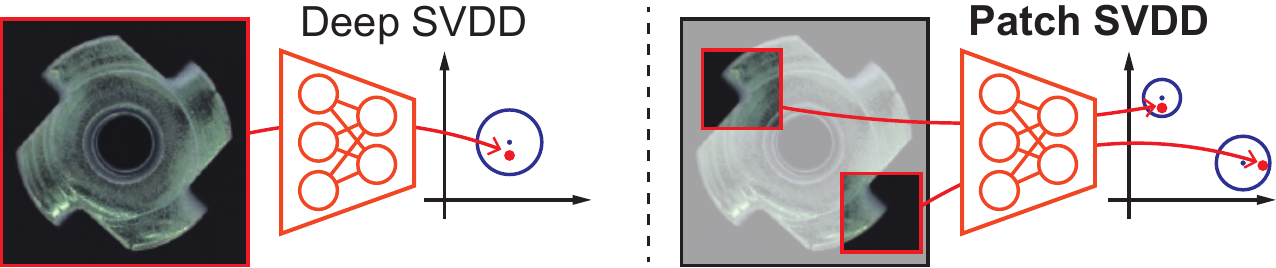}
    \vspace{-1.2em}
    \caption{\textbf{Comparison of Deep SVDD~\cite{deepSVDD} and the proposed method.} Patch SVDD performs inspection on every patch to localize a defect. In addition, the self-supervised learning allows the features to form multi-modal clusters, thereby enhancing anomaly detection capability. The image is from MVTec AD~\cite{mvtecad} dataset.}
    \label{fig:svdd_comparison}
    \vspace{-0.8em}
\end{figure}

\subsubsection{Classifier-based methods} 
After the work of Golan et al.~\cite{geom}, discriminative approaches have been proposed for anomaly detection.
These methods exploit an observation that classifiers lose their confidence~\cite{oodbaseline} for the abnormal input images.
Given an unlabeled dataset, a classifier is trained to predict synthetic labels.
For example, Golan et al.~\cite{geom} randomly flip, rotate, and translate an image, and the classifier is trained to predict the particular type of transformation performed.
If the classifier does not provide a confident and correct prediction, the input image is deemed to be abnormal.
Wang et al.~\cite{e3outlier} proved that such an approach could be extended to an unsupervised scenario, where the training data also contains a few anomalies.
Bergman et al.~\cite{goad} adopted an open-set classification method and generalized the method to include non-image data.

\subsubsection{SVDD-based methods} 
SVDD~\cite{svdd} is a classic one-class classification algorithm.
It maps all the normal training data into a predefined kernel space and seeks the smallest hypersphere that encloses the data in the kernel space.
The anomalies are expected to be located outside the learned hypersphere.
As a kernel function determines the kernel space, the training procedure is merely deciding the radius and center of the hypersphere.

Ruff et al.~\cite{deepSVDD} improved this approach using a deep neural network.
They adopted the neural network in place of the kernel function and trained it along with the radius of the hypersphere.
This modification allows the encoder to learn a data-dependent transformation, thus enhancing detection performance on high-dimensional and structured data.
To avoid a trivial solution (i.e., the encoder outputs a constant), they removed the bias terms in the network.
Ruff et al.~\cite{deep_sad} further applied this method to a semi-supervised scenario.

\begin{figure}[t]
    \centering
    \includegraphics[width=\linewidth]{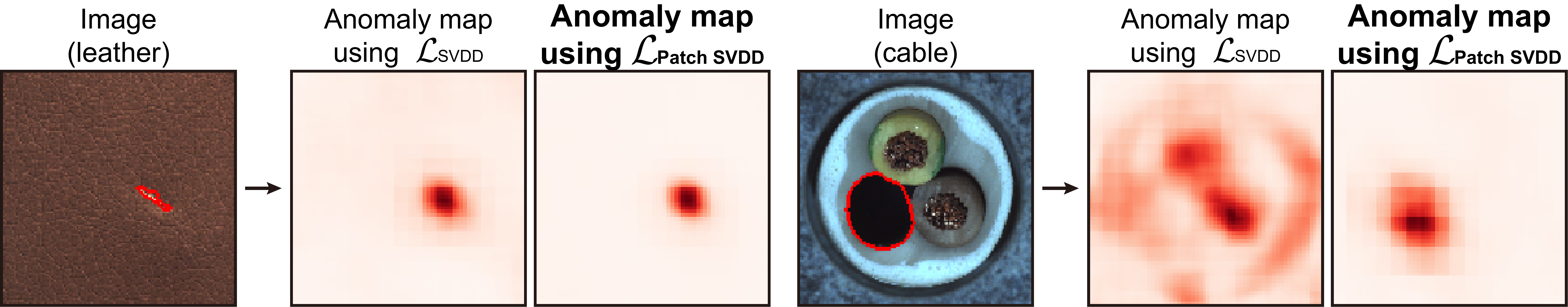}
    \caption{
    \textbf{Comparison of anomaly maps generated using two different losses.}
    For a relatively simple image (leather), the encoders trained with either $\Loss_{\text{SVDD}}$ or $\Loss_{\text{Patch SVDD}}$ both localize the defect (contoured with a red line) well. By contrast, when the image has high complexity (cable), $\Loss_{\text{SVDD}}$ fails to localize the defect. The image is from MVTec AD~\cite{mvtecad} dataset.
    }
    \label{fig:ablation_loss_map}
    \vspace{-5pt}
\end{figure}

\subsection{Self-supervised representation learning}
Learning a representation of an image is a core problem of computer vision.
A series of methods have been proposed to learn a representation of an image without annotation.
One branch of research suggests training the encoder by learning with a \textit{pretext task}, which is a self-labeled task to provide synthetic learning signals.
When a network is trained to solve the pretext task well, the network is expected to be able to extract useful features.
The pretext tasks include predicting relative patch location~\cite{patch_location}, solving a jigsaw puzzle~\cite{jigsaw}, colorizing images~\cite{colorization}, counting objects~\cite{count}, and predicting rotations~\cite{rotation}.

\section{Methods}
\subsection{Patch-wise Deep SVDD} \label{sec:patch_svdd}

Deep SVDD~\cite{deepSVDD} trains an encoder that maps the entire training data to features lying within a small hypersphere in the feature space.
The encoder, $f_\theta$, is trained to minimize the Euclidean distances between the features and the center of the hypersphere using the following loss function:

\begin{equation} \label{eq:deep_svdd}
    \Loss_{\text{SVDD}} = \sum_i \left \| f_\theta (\xvec_i) - \mathbf{c} \right \| _2,
\end{equation}
where $\xvec$ is an input image.
At test time, the distance between the representation of the input and the center is used as an anomaly score.
The center $\mathbf{c}$ is calculated in advance of the training, as shown in Eq.~\ref{eq:deep_svdd_center}, where $N$ denotes the number of the training data.
Therefore, the training pushes the features around a single center.

\begin{equation} \label{eq:deep_svdd_center}
    \mathbf{c} \doteq \frac{1}{N} \sum_i^N  f_\theta (\xvec_i).
\end{equation}

In this study, we extend this approach to patches; the encoder encodes each patch, not the entire image, as illustrated in Fig.~\ref{fig:svdd_comparison}.
Accordingly, inspection is performed for each patch.
Patch-wise inspection has several advantages.
First, the inspection result is available at each position, and hence we can localize the positions of defects.
Second, such fine-grained examination improves overall detection performance.

A direct extension of Deep SVDD~\cite{deepSVDD} to a patch-wise inspection is straightforward.
A patch encoder, $f_\theta$, is trained using $\Loss_{\text{SVDD}}$ with $\xvec$ replaced with a patch, $\pvec$.
The anomaly score is defined accordingly, and the examples of the resulting anomaly maps are provided in Fig.~\ref{fig:ablation_loss_map}.
Unfortunately, the detection performance is poor for the images with high complexity.
This is because patches have high intra-class variation; some patches correspond to the background, while the others contain the object.
As a result, mapping all the features of dissimilar patches to a single center and imposing a uni-modal cluster weaken the connection between the representation and the content.
Therefore, using a single center $\mathbf{c}$ is inappropriate, yet deciding on the appropriate number of multiple centers and the allocation of patches to each center are cumbersome.

\begin{figure}[t]
    \centering
    \includegraphics[width=\textwidth]{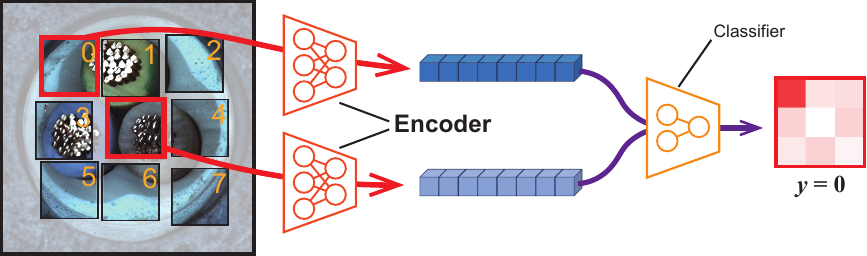}
    \caption{\textbf{Self-supervised learning~\cite{patch_location}.} The encoder is trained to extract informative features so that the following classifier can correctly predict the relative positions of the patches. Once the training is complete, the classifier is discarded. Note that the two encoders share their weights, as in Siamese Network~\cite{siamese}. The image is from MVTec AD~\cite{mvtecad} dataset.}
    \label{fig:self_supervision}
    \vspace{-0.5em}
\end{figure}

To bypass the above issues, we do not explicitly define the center and allocate the patches.
Instead, we train the encoder to gather semantically similar patches by itself.
The semantically similar patches are obtained by sampling spatially adjacent patches, and the encoder is trained to minimize the distances between their features using the following loss function:

\begin{equation} \label{eq:patch_svdd}
    \Loss_{\text{SVDD'}} = \sum_{i,i'} \left \| f_\theta (\pvec_i) - f_\theta (\pvec_{i'}) \right \| _2,
\end{equation}
where $\pvec_{i'}$ is a patch near $\pvec_{i}$.
Furthermore, to enforce the representation to capture the semantics of the patch, we append the following self-supervised learning.

\subsection{Self-supervised learning} \label{sec:self_supervised}
Doersch et al.~\cite{patch_location} trained an encoder and classifier pair to predict the relative positions of two patches, as depicted in Fig.~\ref{fig:self_supervision}.
A well-performing pair implies that the trained encoder extracts useful features for location prediction.
Aside from this particular task, previous research~\cite{jigsaw,rotation,revisit_ssl} reported that the self-supervised encoder functions as a powerful feature extractor for downstream tasks.

For a randomly sampled patch $\pvec_1$, Doersch et al.~\cite{patch_location} sampled another patch $\pvec_2$ from one of its eight neighborhoods in a 3 $\times$ 3 grid.
If we let the true relative position be $y \in \{0, ..., 7\}$, the classifier $C_{\phi}$ is trained to predict $y=C_\phi(f_\theta(\pvec_1), f_\theta(\pvec_2))$ correctly.
The size of the patch is the same as the receptive field of the encoder.
To prevent the classifier from exploiting shortcuts~\cite{lens} (e.g., color aberration), we randomly perturb the RGB channels of the patches.
Following the approach by Doersch et al.~\cite{patch_location}, we add a self-supervised learning signal by adding the following loss term:
\begin{equation} \label{eq:ssl}
    \Loss_{\text{SSL}} = \texttt{Cross-entropy} \left (y, C_\phi \left ( f_\theta(\pvec_1), f_\theta(\pvec_2) \right ) \right ).
\end{equation}
As a result, the encoder is trained using a combination of two losses with the scaling hyperparameter $\lambda$, as presented in Eq.~\ref{eq:patch_svdd_final}.
This optimization is performed using stochastic gradient descent and Adam optimizer~\cite{adam}.

\begin{equation} \label{eq:patch_svdd_final}
    \Loss_{\text{Patch SVDD}} = \lambda \Loss_{\text{SVDD'}} + \Loss_{\text{SSL}}.
\end{equation}

\begin{figure}[t]
    \centering
    \includegraphics[width=\textwidth]{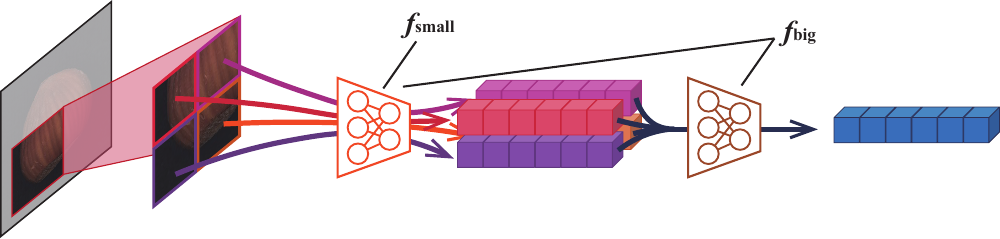}
    \caption{\textbf{Hierarchical encoding.} An input patch is divided into a 2 $\times$ 2 grid of sub-patches, and the sub-patches are independently encoded using the smaller encoder ($f_{\text{small}}$). The output features are aggregated to produce a single feature. The image is from MVTec AD~\cite{mvtecad} dataset.}
    \label{fig:hierarchical}
\end{figure}
\subsection{Hierarchical encoding} \label{sec:hierarchical}
As anomalies vary in size, deploying multiple encoders with various receptive fields helps in dealing with variation in size.
The experimental results in Section~\ref{sec:hierarchical_results} show that enforcing a hierarchical structure on the encoder boosts anomaly detection performance as well.
For this reason, we employ a hierarchical encoder that embodies a smaller encoder; the hierarchical encoder is defined as
\begin{equation} \label{eq:hierarchical}
    f_{\text{big}}(\pvec)=g_{\text{big}}(f_{\text{small}}(\pvec)).
\end{equation}

An input patch $\pvec$ is divided into a 2 $\times$ 2 grid, and their features are aggregated to constitute the feature of $\pvec$, as shown in Fig.~\ref{fig:hierarchical}.
Each encoder with receptive field size $K$ is trained with the self-supervised task of patch size $K$.
Throughout the experiment, the receptive field sizes of the large and small encoders are 64 and 32, respectively.


\begin{figure}[t]
    \centering
    \includegraphics[width=\textwidth]{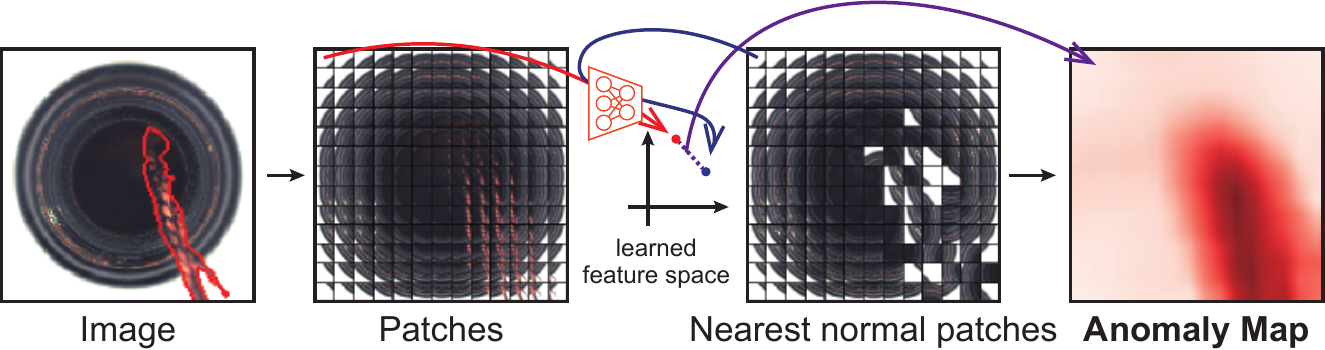}
    \vspace{-1em}
    \caption{\textbf{Overall flow of the proposed method.} For a given test image, Patch SVDD divides the image into patches of size $K$ with strides $S$ and extracts their features using the trained encoder. The L2 distance to the nearest normal patch in the feature space becomes the anomaly score of each patch. The resulting anomaly map localizes the defects (contoured with a red line). The image is from MVTec AD~\cite{mvtecad} dataset.}
    \vspace{-1em}
    \label{fig:overall_method}
\end{figure}

\subsection{Generating anomaly maps}
After training the encoders, the representations from the encoder are used to detect the anomalies.
First, the representation of every normal train patch, $\left \{f_\theta (\pvec_{\text{normal}}) | \pvec_{\text{normal}}  \right \}$, is calculated and stored.
Given a query image $\xvec$, for every patch $\pvec$ with a stride $S$ within $\xvec$, the L2 distance to the nearest normal patch in the feature space is then defined to be its anomaly score using Eq.~\ref{eq:anomaly_score_patch}.
To mitigate the computational cost of the nearest neighbor search, we adopt its approximate algorithm\footnote{\url{https://github.com/yahoojapan/NGT}}.
As a result, the inspection of a single image from MVTec AD~\cite{mvtecad} dataset for example, requires approximately 0.48 second.

\begin{equation} \label{eq:anomaly_score_patch}
    \Atheta^{\text{patch}}(\pvec) \doteq \min_{\pvec_{\text{normal}}} \left \| f_\theta (\pvec) - f_\theta (\pvec_{\text{normal}}) \right \| _2.
\end{equation}

Patch-wise calculated anomaly scores are then distributed to the pixels.
As a consequence, pixels receive the average anomaly scores of every patch to which they belong, and we denote the resulting anomaly map as $\mmap$.

The multiple encoders discussed in Section~\ref{sec:hierarchical} constitute multiple feature spaces, thereby yielding multiple anomaly maps.
We aggregate the multiple maps using element-wise multiplication, and the resulting anomaly map, $\mmap_{\text{multi}}$, provides the answer to the problem of anomaly segmentation:

\begin{equation} \label{eq:anomaly_map_multi}
    \mmap_{\text{multi}} \doteq \mmap_{\text{small}} \odot \mmap_{\text{big}},
\end{equation}
where $\mmap_{\text{small}}$ and $\mmap_{\text{big}}$ are the generated anomaly maps using $f_{\text{small}}$ and $f_{\text{big}}$, respectively.
The pixels with high anomaly scores in the map $\mmap_{\text{multi}}$ are deemed to contain defects.

It is straightforward to address the problem of anomaly detection.
The maximum anomaly score of the pixels in an image is its anomaly score, as expressed in Eq.~\ref{eq:anomaly_score_image}.
Fig.~\ref{fig:overall_method} illustrates the overall flow of the proposed method, and its pseudo-code is provided in Appendix~\ref{sec:appendix_pseudo_code}.

\begin{equation} \label{eq:anomaly_score_image}
    \Atheta^{\text{image}}(\xvec) \doteq \max_{i,j} \mmap_{\text{multi}}(\xvec)_{ij}.
\end{equation}

\begin{figure}[t]
    \centering
    \includegraphics[width=\linewidth]{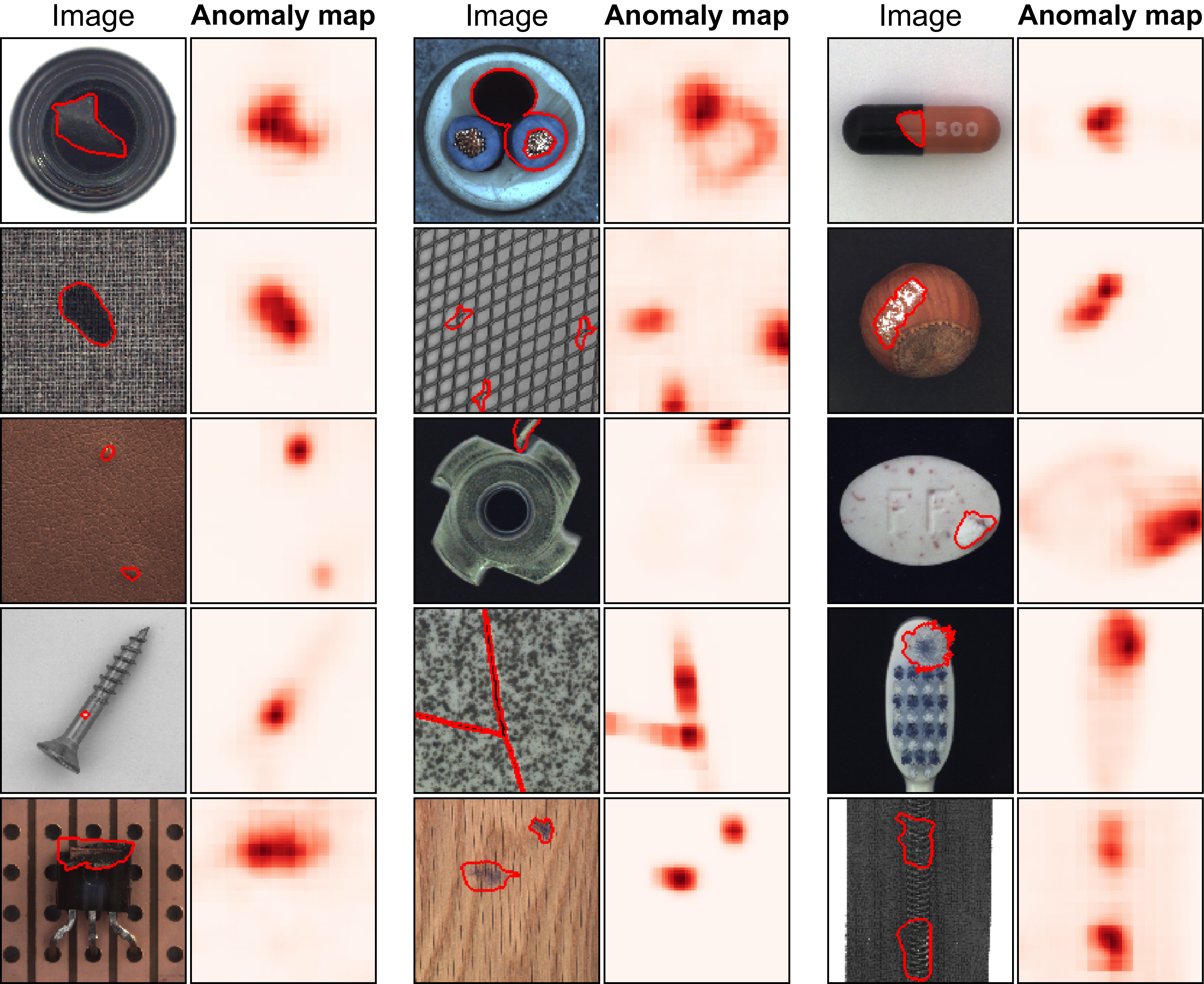}
    \vspace{-10pt}
    \caption{\textbf{Anomaly maps generated by the proposed method.} Patch SVDD generates anomaly maps of each image in fifteen classes of MVTec AD~\cite{mvtecad} dataset. The ground truth defect annotations are depicted as red contours in the image, and the darker heatmap indicates higher anomaly scores.
    }
    \label{fig:anomaly_maps}
    \vspace{-5pt}
\end{figure}

\section{Results and Discussion}
To verify the validity of the proposed method, we applied it to MVTec AD~\cite{mvtecad} dataset.
The dataset consists of 15-class industrial images, each class categorized as either an \textit{object} or \textit{texture}.
Ten \textit{object} classes contain regularly positioned objects, whereas the \textit{texture} classes contain repetitive patterns.
The implementation details used throughout the study are provided in Appendix~\ref{sec:appendix_implementation}, and please refer to \cite{mvtecad} for more details on the dataset.

\begin{table}[t]
  \centering
  \caption{\textbf{Anomaly detection (left) and segmentation (right) performances on MVTec AD~\cite{mvtecad} dataset.} The proposed method, Patch SVDD, achieves the state-of-the-art performances on both tasks.}
  \vspace{0.5em}
  \begin{minipage}{0.5\linewidth}
   \centering
   \begin{tabular}{l @{\hskip 0.07in}c}
    \Xhline{1pt}\\[-0.95em]
     Method     & AUROC\\
    \hline\hline
    \\[-0.9em]
    \multicolumn{2}{l}{Task: Anomaly Detection}                     \\
    $\text{Deep SVDD~\cite{deepSVDD}}_{\text{~~ICML' 18}}$  & 0.592 \\
    $\text{GEOM~\cite{geom}}_{\text{~~NeurIPS' 18}}$        & 0.672\\
    $\text{GANomaly~\cite{ganomaly}}_{\text{~~ACCV' 18}}$   & 0.762\\
    $\text{ITAE~\cite{itae}}_{\text{~~arXiv' 19}}$          & 0.839 \\
    \textbf{Patch SVDD (Ours)}                              & \textbf{0.921} \\
    \\[-0.9em]
    \Xhline{1pt}
    \end{tabular}%
  \end{minipage}\hfill
  \begin{minipage}{0.5\linewidth}
   \centering
   \begin{tabular}{l @{\hskip 0.07in}c}
    \Xhline{1pt}\\[-0.95em]
     Method     & AUROC\\
    \hline\hline
    \\[-0.9em]
    \multicolumn{2}{l}{Task: Anomaly Segmentation}                          \\
    $\text{L2-AE}$                                                  & 0.804 \\
    $\text{SSIM-AE}$                                                & 0.818 \\
    $\text{VE VAE}_{\text{~~CVPR' 20}}$~\cite{ve_vae}               & 0.861 \\
    $\text{VAE Proj}_{\text{~~ICLR' 20}}$~\cite{iterative_project}  & 0.893 \\
    \textbf{Patch SVDD (Ours)}                      & \textbf{0.957} \\
    \\[-0.9em]
    \Xhline{1pt}
    \end{tabular}%
  \end{minipage}\hfill
  \vspace{-1em}
  \label{table:anomaly_det_seg}
\end{table}

\subsection{Anomaly detection and segmentation results}
Fig.~\ref{fig:anomaly_maps} shows anomaly maps generated using the proposed method, indicating that the defects are properly localized, regardless of their size.
Table~\ref{table:anomaly_det_seg} shows the detection and segmentation performances for MVTec AD~\cite{mvtecad} dataset compared with state-of-the-art baselines in AUROC.
Patch SVDD provides state-of-the-art performance over the powerful baselines including auto encoder-based and classifier-based methods and outperforms Deep SVDD~\cite{deepSVDD} by 55.6\% improvement.
More numerical results are provided in Appendix~\ref{sec:appendix_numerical}.

\begin{figure}[t]
    \centering
    \includegraphics[width=\linewidth]{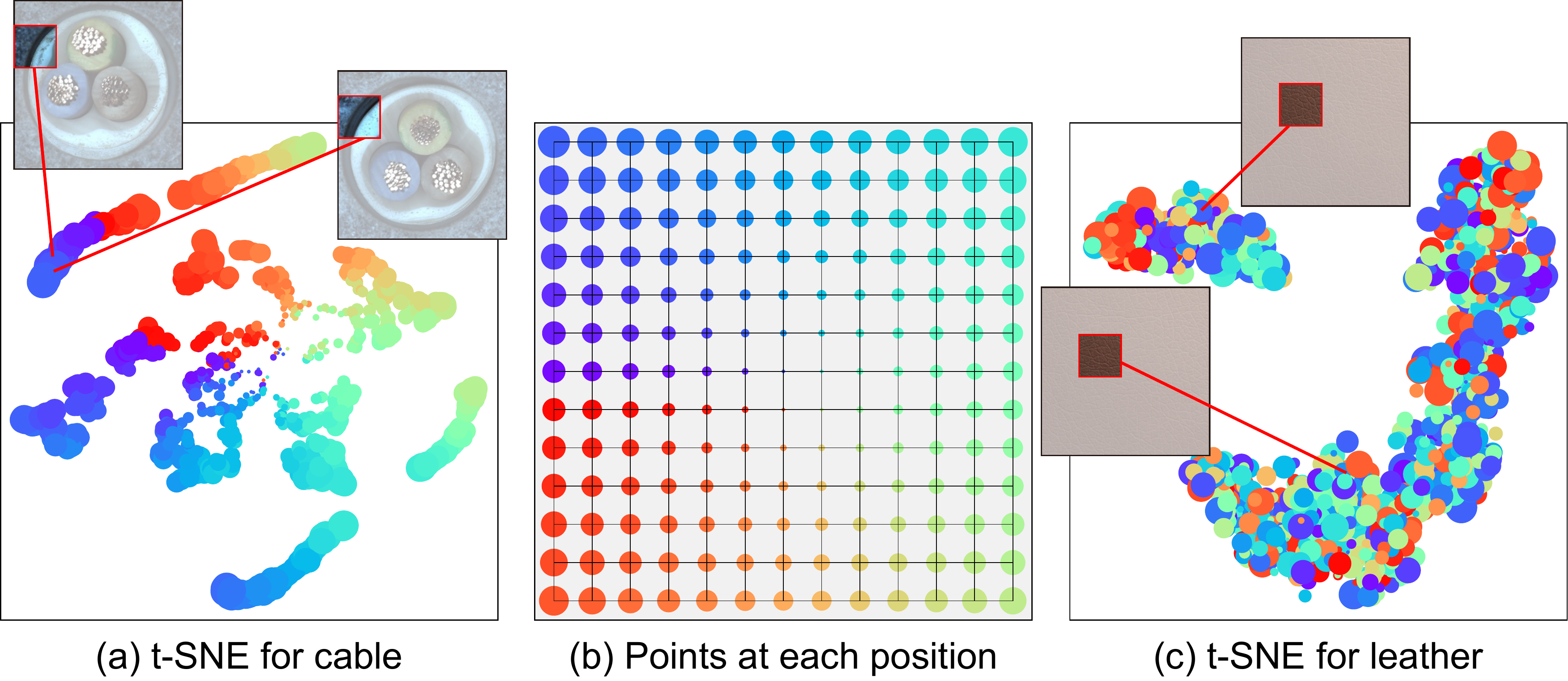}
    \caption{\textbf{t-SNE visualizations~\cite{tsne} of the learned features.} The color and size of each point represent the position ($\theta$ and $r$ of the polar coordinates) within an image (b). From its color and size, we can infer the positions of the corresponding patches of the features in (a, c).
    }
    \label{fig:tsne}
    \vspace{-5pt}
\end{figure}

\subsection{Detailed analysis}
\subsubsection{t-SNE visualization}
Fig.~\ref{fig:tsne} shows t-SNE visualizations~\cite{tsne} of the learned features of multiple train images.
Patches located at the points shown in Fig.~\ref{fig:tsne}(b) are mapped to the points with the same color and size in Fig.~\ref{fig:tsne}(a) and Fig.~\ref{fig:tsne}(c).
In Fig.~\ref{fig:tsne}(a), the points with similar color and size form clusters in the feature space.
Since the images in the cable class are regularly positioned, the patches from the same position have similar content, even if they are from different images.
Likewise, for the regularly positioned \textit{object} classes, the points with similar color and size in t-SNE visualization (i.e., the patches with similar positions) can be regarded to be semantically similar.
By contrast, features of the leather class in Fig.~\ref{fig:tsne}(c) show the opposite tendency.
This is because the patches in \textit{texture} classes are analogous, regardless of their position in the image; the positions of the patches are not quite related to their semantics for the \textit{texture} images.

\setlength{\tabcolsep}{0.03in}
\begin{table}[t]
  \begin{minipage}{0.43\linewidth}
    \vspace{-1em}
    \center
    \caption{\textbf{The effect of the losses.} Modifying $\Loss_{\text{SVDD}}$ to $\Loss_{\text{SVDD'}}$ and adopting $\Loss_{\text{SSL}}$ both improve the anomaly detection ($\texttt{Det.}$) and segmentation ($\texttt{Seg.}$) performances.}
    \vspace{0.5em}
    \begin{tabular}{c c c |c c }  
         \Xhline{1pt}
         \\[-0.9em]
        $\Loss_{\text{SVDD}}$ & $\Loss_{\text{SVDD'}}$ & $\Loss_{\text{SSL}}$ &  \texttt{Det.}   & \texttt{Seg.} \\
        \hline\hline
        \\[-0.9em]
        \textred{\cmark}   & \xmark     &   \xmark    & 0.703     &  0.832   \\
        \xmark   & \textred{\cmark}     &   \xmark    & 0.739     &  0.880   \\
        \xmark   & \textred{\cmark}     &   \textred{\cmark}    & \textbf{0.921}     &  \textbf{0.957}    \\
        \Xhline{1pt}
     \end{tabular}
     \vspace{1em}
    \label{table:ablation_loss}
  \end{minipage}\hfill
  \begin{minipage}{0.55\linewidth}
    \centering
    \vspace{0.1em}
    \includegraphics[width=\linewidth]{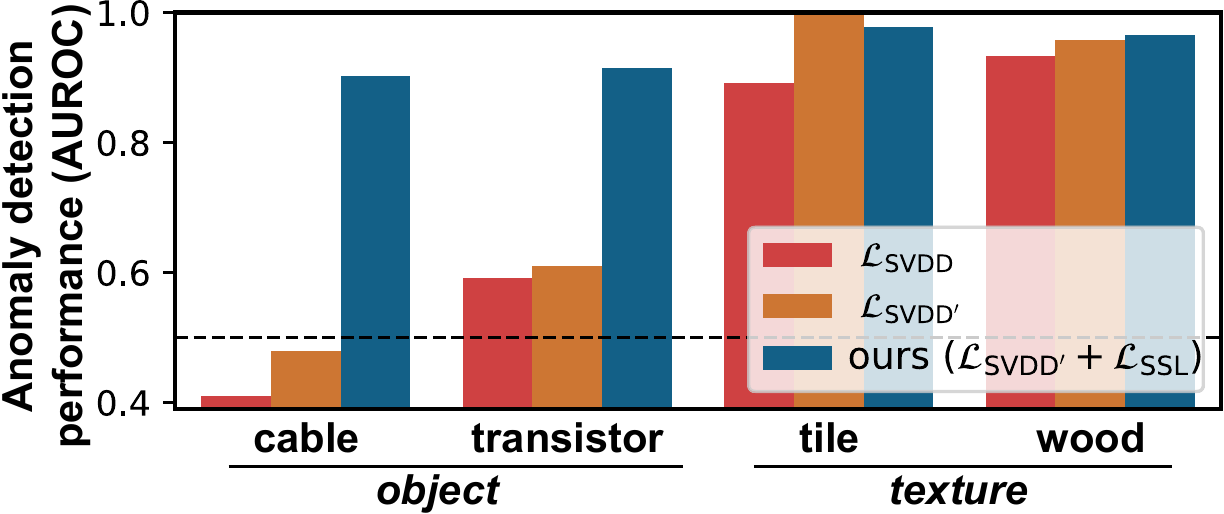}
    \vspace{-0.5em}
    \captionof{figure}{\label{fig:ablation_loss} \textbf{The effects of the losses vary among classes.} $\Loss_{\text{SSL}}$ is particularly beneficiary to the \textit{object} classes.}
  \end{minipage}\hfill
\end{table}

\begin{figure}[t]
  \centering
  \includegraphics[width=1\linewidth]{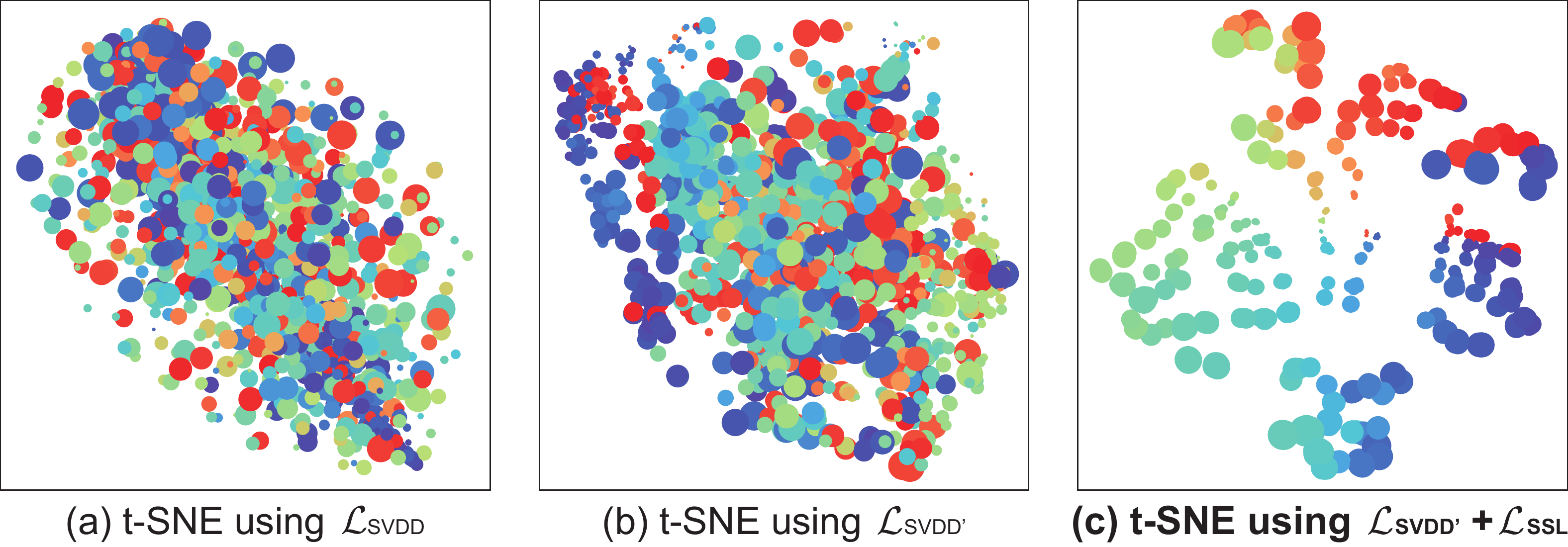}
  \vspace{-1.3em}
  \captionof{figure}{\label{fig:ablation_loss_tsne} \textbf{Features of the multiple train images in the transistor class by the encoders trained with different losses}.
   Adopting $\Loss_{\text{SSL}}$ (c) enables the representations to form clusters on the basis of their semantics.}
  \vspace{-1em}
\end{figure}

\subsubsection{Effect of self-supervised learning} \label{sec:ssl_effect}
Patch SVDD trains an encoder using two losses: $\Loss_{\text{SVDD'}}$ and $\Loss_{\text{SSL}}$, where $\Loss_{\text{SVDD'}}$ is a variant of $\Loss_{\text{SVDD}}$.
To compare the roles of the proposed loss terms, we conduct an ablation study.
Table~\ref{table:ablation_loss} suggests that the modification of $\Loss_{\text{SVDD}}$ to $\Loss_{\text{SVDD'}}$ and the adoption of $\Loss_{\text{SSL}}$ improve the anomaly detection and segmentation performances.
Fig.~\ref{fig:ablation_loss} shows that the effects of the proposed loss terms vary among classes.
Specifically, the \textit{texture} classes (e.g. tile and wood) are less sensitive to the choice of loss, whereas the \textit{object} classes, including cable and transistor, benefit significantly from $\Loss_{\text{SSL}}$.

To investigate the reason behind these observations, we provide (in Fig.~\ref{fig:ablation_loss_tsne}) t-SNE visualizations of the features of an \textit{object} class (the transistor) for the encoders trained with $\Loss_{\text{SVDD}}$, $\Loss_{\text{SVDD'}}$, and $\Loss_{\text{SVDD'}} + \Loss_{\text{SSL}}$.
When training is performed with $\Loss_{\text{SVDD}}$ (Fig.~\ref{fig:ablation_loss_tsne}(a)) or $\Loss_{\text{SVDD'}}$ (Fig.~\ref{fig:ablation_loss_tsne}(b)), the features form a uni-modal cluster.
In contrast, $\Loss_{\text{SSL}}$ results in multi-modal feature clusters on the basis of their semantics (i.e., color and size), as shown in Fig.~\ref{fig:ablation_loss_tsne}(c).
The multi-modal property of the features is particularly beneficial to the \textit{object} classes, which have high intra-class variation among the patches.
Features of the patches with dissimilar semantics are separated, and hence anomaly inspection using those features becomes more deliberate and accurate.

\begin{figure}[t]
  \begin{minipage}{0.32\linewidth}
    \centering
    \vspace{-1.5em}
    \includegraphics[width=\linewidth]{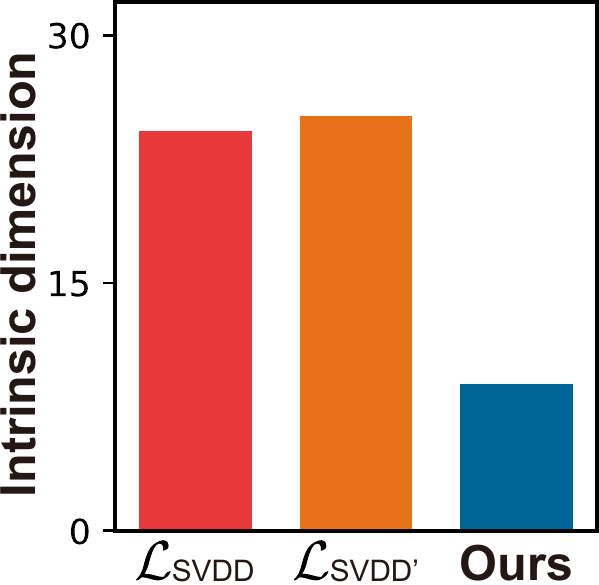}
    \vspace{-0.5em}
    \caption{\textbf{Intrinsic dimensions of the features under different losses}.}
    \label{fig:ablation_loss_id}
  \end{minipage}\hfill
  \begin{minipage}{0.63\linewidth}
    \centering
    \includegraphics[width=\linewidth]{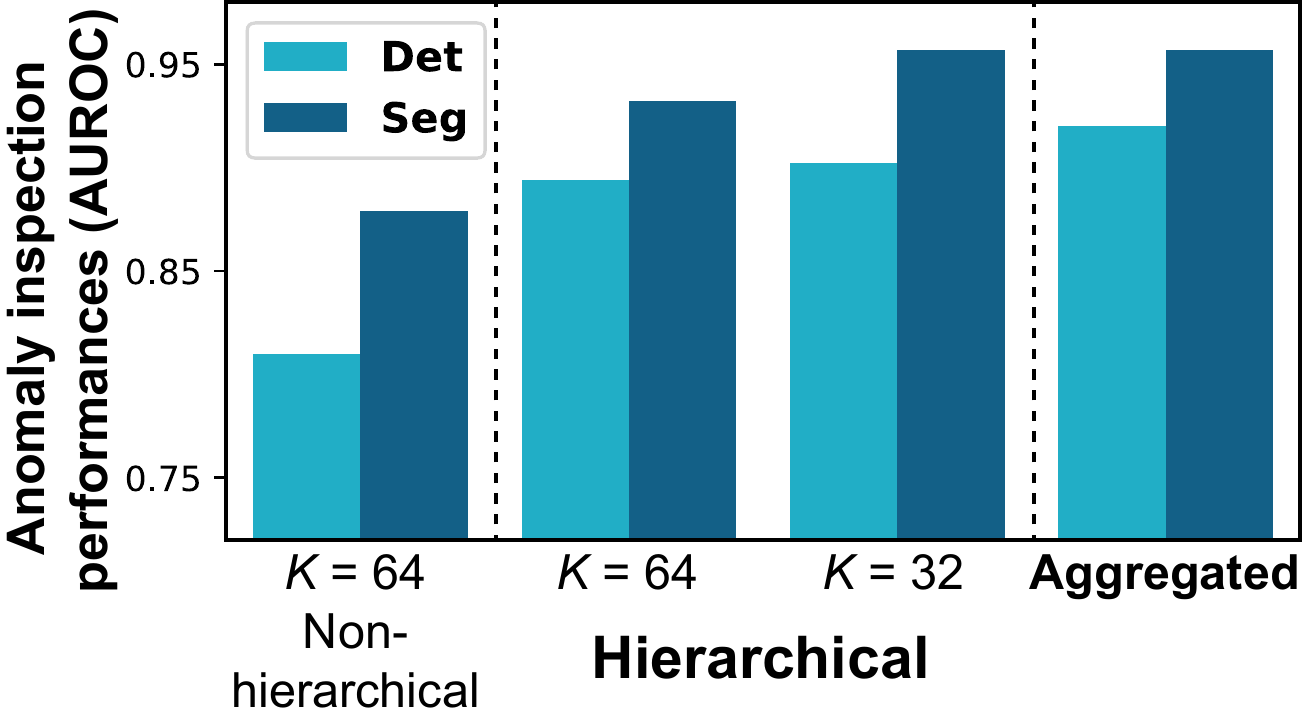}
    \vspace{-1.5em}
    \caption{\textbf{The effect of hierarchical encoding.} Aggregating the results from multi-scale inspection boosts the performance, and adopting hierarchical structure to the encoder is helpful as well.
    }
  \label{fig:hierarchical_helps}
  \end{minipage}\hfill
  \vspace{-1.5em}
\end{figure}

The intrinsic dimensions (ID)~\cite{intrinsic_dimension} of the features also indicate the effectiveness of $\Loss_{\text{SSL}}$.
The ID is the minimal number of coordinates required to describe the points without significant information loss~\cite{intrinsic_dimension2}.
A larger ID denotes that the points are spreaded in every direction, while a smaller ID indicates that the points lie on low-dimensional manifolds with high separability.
In Fig.~\ref{fig:ablation_loss_id}, we show the average IDs of features in each class trained with three different losses.
If the encoder is trained with the proposed $\Loss_{\text{Patch SVDD}}$, features with the lowest ID are yielded, implying that these features are neatly distributed.

\subsubsection{Hierarchical encoding} \label{sec:hierarchical_results}
In Section~\ref{sec:hierarchical}, we proposed the use of hierarchical encoders.
Fig.~\ref{fig:hierarchical_helps} shows that aggregating multi-scale results from multiple encoders improves the inspection performances.
In addition, an ablation study with a non-hierarchical encoder shows that the hierarchical structure itself also boosts performance.
We postulate that the hierarchical architecture provides regularization for the feature extraction.
Note that the non-hierarchical encoder has a number of parameters similar to that of the hierarchical counterpart.

\begin{figure}[t]
    \centering
    \includegraphics[width=\linewidth]{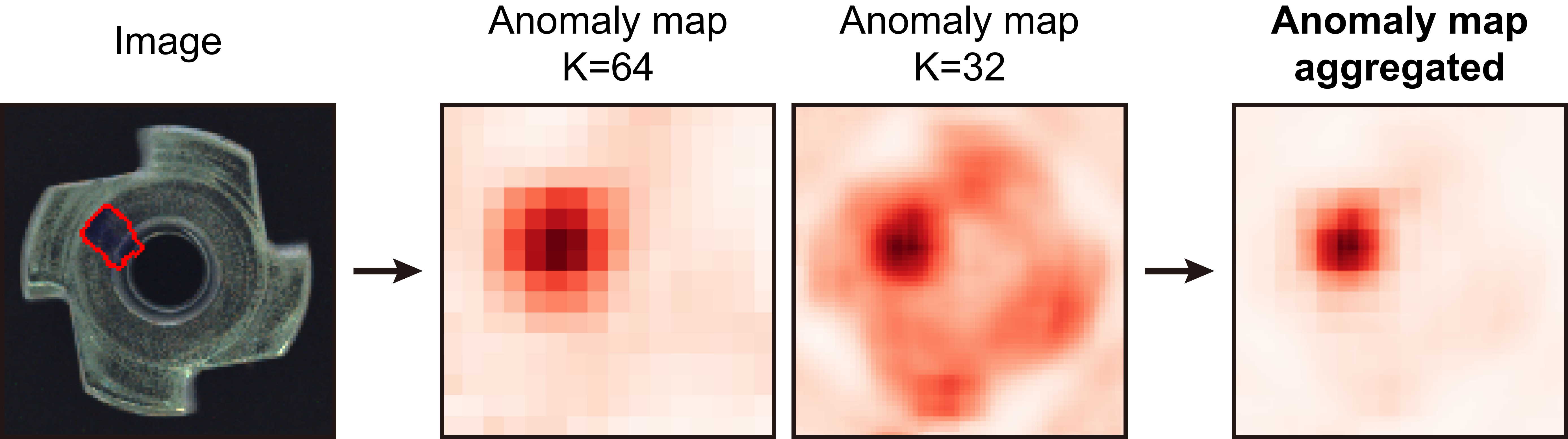}
    \vspace{-10pt}
    \caption{\textbf{Multi-scale inspection.} Patch SVDD performs multi-scale inspection and aggregates the results. The image is from MVTec AD~\cite{mvtecad} dataset.
    }
    \label{fig:hierarchical_maps}
    \vspace{-5pt}
\end{figure}

We provide an example of multi-scale inspection results, together with an aggregated anomaly map, in Fig.~\ref{fig:hierarchical_maps}.
The anomaly maps from various scales provide complementary inspection results; the encoder with a large receptive field coarsely locates the defect, whereas the one with a smaller receptive field refines the result.
Therefore, an element-wise multiplication of the two maps localizes the accurate position of the defect.

\subsubsection{Hyperparameters}
\begin{figure}[t]
  \centering
  \begin{minipage}{0.32\linewidth}
  \centering
    \vspace{1.2em}
    \includegraphics[width=\linewidth]{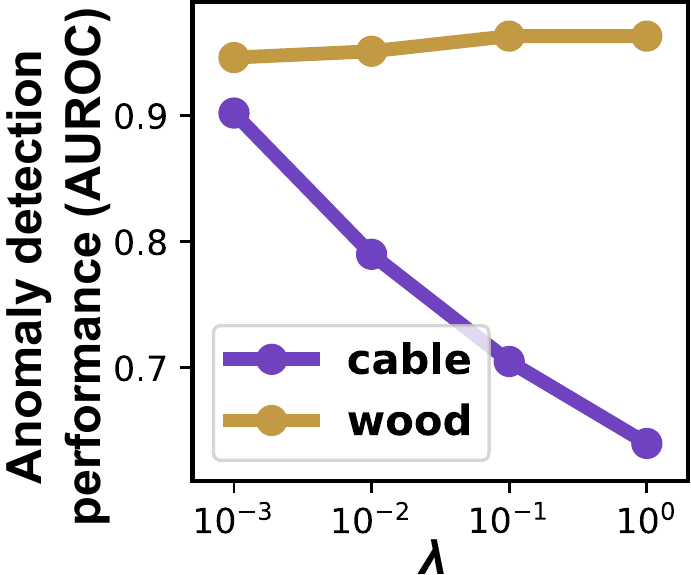}
    \vspace{-2em}
    \caption{\textbf{The effect of $\lambda$.} The anomaly detection performances for the two classes show the opposite trends as $\lambda$ varies.}
    \label{fig:ablation_lambda} 
  \end{minipage}\hfill
  \begin{minipage}{0.63\linewidth}
    \centering
    \vspace{-0.3em}
    \includegraphics[width=\linewidth]{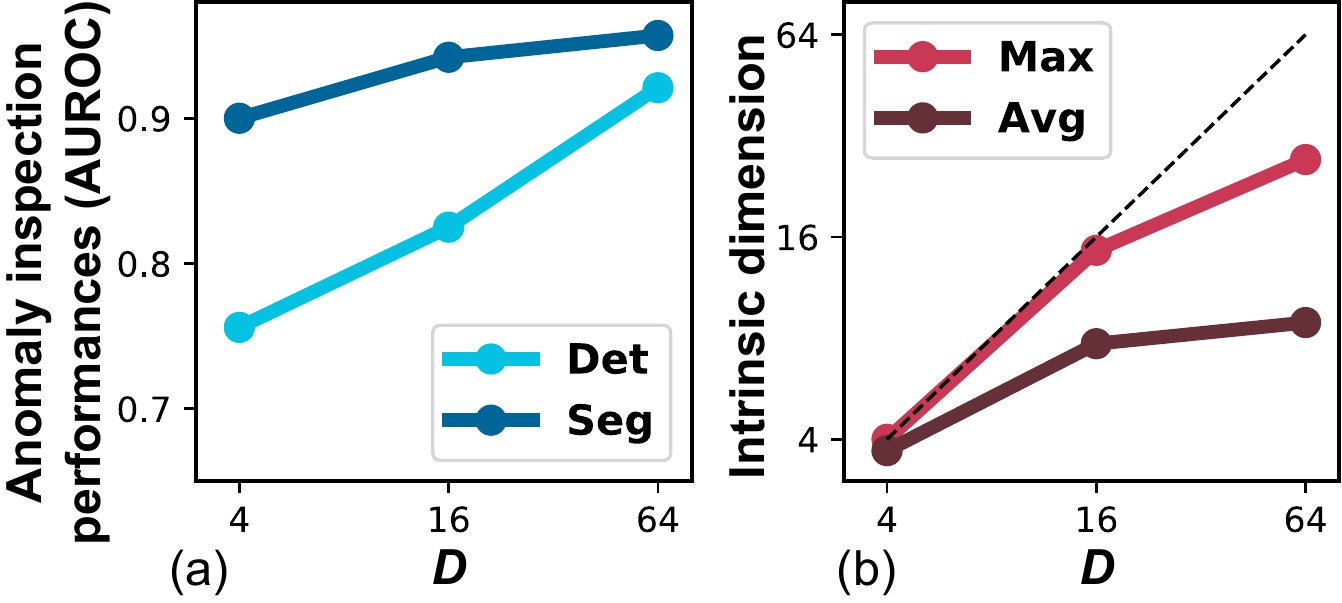}
    \vspace{-1em}
    \caption{\textbf{The effect of the embedding dimension, $D$.} Larger $D$ yields better inspection results (a) and larger intrinsic dimensions (b).}
    \label{fig:ablation_D}
  \end{minipage}\hfill
  \vspace{-1em}
\end{figure}

As shown in Eq.~\ref{eq:patch_svdd_final}, the hyperparameter $\lambda$ balances $\Loss_{\text{SVDD'}}$ and $\Loss_{\text{SSL}}$.
A large $\lambda$ emphasizes gathering of the features, while a small $\lambda$ promotes their informativeness.
Interestingly, the most favorable value of $\lambda$ varies among the classes.
Anomalies in the \textit{object} classes are well detected under a smaller $\lambda$, while the \textit{texture} classes are well detected with a larger $\lambda$.
Fig.~\ref{fig:ablation_lambda} shows an example of this difference; the anomaly detection performance for the cable class (\textit{object}) improves as $\lambda$ decreases, while the wood class (\textit{texture}) shows the opposite trend.
As discussed in the previous sections, this occurs because the self-supervised learning is more helpful when the patches show high intra-class variation, which is the case for the \textit{object} classes.
The result coincides with that shown in Fig.~\ref{fig:ablation_loss} because using $\Loss_{\text{SVDD'}}$ as a loss is equivalent to using $\Loss_{\text{Patch SVDD}}$ with $\lambda >> 1$.

The number of feature dimensions, $D$, is another hyperparameter of the encoder.
The anomaly inspection performance for varying $D$ is depicted in Fig.~\ref{fig:ablation_D}(a).
A larger $D$ signifies improved performance---a trend that has been discussed in a self-supervised learning venue~\cite{revisit_ssl}.
Fig.~\ref{fig:ablation_D}(b) indicates that the ID of the resulting features increases with increasing $D$.
The black dashed line represents the $y=x$ graph, and it is the upper bound of ID.
The average ID of features among the classes saturates as $D=64$; therefore, we used a value of $D=64$ throughout our study.

\subsubsection{Random encoder}
\setlength{\tabcolsep}{0.12in}
\begin{table}[t]
  \centering
  \caption{\textbf{Nearest neighbor algorithm using the random encoders and raw patches for MVTec AD~\cite{mvtecad} dataset.} For certain classes, the nearest neighbor algorithm using the random features shows good anomaly detection performance. For those classes, using the raw patches also yields high performance.}
  \vspace{0.5em}
    \begin{tabular}
    {l @{\hskip 0.1in} |c c |c c |c }
    \Xhline{1pt}\\[-0.95em]
    & \multicolumn{2}{c|}{\textbf{Random Encoder}} & \multicolumn{2}{c|}{\textbf{Raw Patch}} & \multicolumn{1}{c}{L2-AE} \\
    Classes      & \texttt{Det.} & \texttt{Seg.} & \texttt{Det.} & \texttt{Seg.} & \texttt{Seg.} \\
     
    \hline\hline
    \\[-0.9em]
    bottle         & 0.89 & 0.80 & \textbf{0.92 } & 0.82 & 0.91 \\
    cable          & 0.56 & 0.70 & 0.57 & 0.84 & 0.73 \\
    capsule        & 0.67 & \textbf{0.93} & 0.67 & \textbf{0.94} & 0.79 \\
    carpet         & 0.42 & 0.66 & 0.48 & 0.74 & 0.54 \\
    grid           & 0.75 & 0.61 & 0.83 & 0.76 & 0.96 \\
    hazelnut       & 0.83 & \textbf{0.93} & 0.83 & \textbf{0.95} & 0.98 \\
    leather        & 0.69 & 0.85 & 0.69 & 0.88 & 0.75 \\
    metal\_nut     & 0.37 & 0.79 & 0.52 & 0.89 & 0.88 \\
    pill           & 0.72 & \textbf{0.92} & 0.77 & \textbf{0.92} & 0.89 \\
    screw          & 0.69 & \textbf{0.93} & 0.56 & \textbf{0.93} & 0.98 \\
    tile           & 0.68 & 0.54 & 0.70 & 0.51 & 0.48 \\
    toothbrush     & 0.81 & \textbf{0.95} & \textbf{0.92} & \textbf{0.97} & 0.97 \\
    transistor     & 0.53 & 0.88 & 0.68 & \textbf{0.92} & 0.91 \\
    wood           & \textbf{0.90} & 0.74 & \textbf{0.94} & 0.78 & 0.63 \\
    zipper         & 0.70 & 0.84 & \textbf{0.94} & \textbf{0.92} & 0.68 \\
    \Xhline{0.5pt}
    \end{tabular}%
\vspace{0.5mm}
  
  \label{table:random_enc}
\end{table}
Doersch et al.~\cite{patch_location} showed that randomly initialized encoders perform reasonably well in image retrieval; given an image, the nearest images in the random feature space look similar to humans as well.
Inspired by this observation, we examined the anomaly detection performance of the random encoders and provided the results in Table~\ref{table:random_enc}.
As in Eq.~\ref{eq:anomaly_score_patch}, the anomaly score is defined to be the distance to the nearest normal patch, but in the random feature space.
In the case of certain classes, the features of the random encoder are effective in distinguishing between normal and abnormal images.
Some results even outperform the trained deep neural network model (L2-AE).

Here, we investigate the reason for the high separability of the random features.
For simplicity, let us assume the encoder to be a one-layered convolutional layer parametrized by a weight $W \neq \mathbf{0}$ and a bias $b$ followed by a nonlinearity, $\sigma$.
Given two patches $\pvec_1$ and $\pvec_2$, their features $h_1$ and $h_2$ are provided by Eq.~\ref{eq:random_encoder_define}, where $*$ denotes a convolution operation.

\begin{equation} \label{eq:random_encoder_define}
\begin{split}
    & h_1 = \sigma (W * \pvec_1 + b) \\
    & h_2 = \sigma (W * \pvec_2 + b).
\end{split}
\end{equation}

As suggested by Eq.~\ref{eq:random_encoder}, when the features are close, so are the patches, and vice versa.
Therefore, retrieving the nearest patch in the feature space is analogous to doing so in the image space.

\begin{equation} \label{eq:random_encoder}
\begin{split}
    \left \| h_1 - h_2 \right \| _2 \approx 0 & \Leftrightarrow   (W * \pvec_1 + b) - (W * \pvec_2 + b) \approx 0  \\
    & \Leftrightarrow  W * (\pvec_1 - \pvec_2) \approx 0 \\
    & \Leftrightarrow  \left \| \pvec_1 - \pvec_2 \right \| _2 \approx 0.
\end{split}
\end{equation}

In Table~\ref{table:random_enc}, we also provide the results for anomaly detection task using the nearest neighbor algorithm using the raw patches (i.e., $f_\theta(\pvec)=\pvec$ in Eq.~\ref{eq:anomaly_score_patch}).
For certain classes, the raw patch nearest neighbor algorithm works surprisingly well.
The effectiveness of the raw patches for anomaly detection can be attributed to the high similarity among the normal images.

Furthermore, the well-separated classes provided by the random encoder are well-separated by the raw patch nearest neighbor algorithm, and vice versa.
Together with the conclusion of Eq.~\ref{eq:random_encoder}, this observation implies the strong relationship between the raw image patch and its random feature.
To summarize, the random features of anomalies are easily separable because they are alike the raw patches, and the raw patches are easily separable.

\section{Conclusion}

In this paper, we proposed Patch SVDD, a method for image anomaly detection and segmentation.
Unlike Deep SVDD~\cite{deepSVDD}, we inspect the image at the patch level, and hence we can also localize defects.
Moreover, additional self-supervised learning improves detection performance.
As a result, the proposed method achieved state-of-the-art performance on MVTec AD~\cite{mvtecad} industrial anomaly detection dataset.

In previous studies~\cite{jigsaw,colorization}, images were featurized prior to the subsequent downstream tasks because of their high-dimensional and structured nature.
However, the results in our analysis suggest that a nearest neighbor algorithm with a raw patch often discriminates anomalies surprisingly well.
Moreover, since the distances in random feature space are closely related to those in the raw image space, random features can provide distinguishable signals.


\bibliographystyle{splncs}
\bibliography{references}

\clearpage
\appendix

\setcounter{section}{0}
\renewcommand\thesection{A\arabic{section}}
\setcounter{table}{0}
\renewcommand{\thetable}{A\arabic{table}}
\setcounter{figure}{0}
\renewcommand{\thefigure}{A\arabic{figure}}
\setcounter{equation}{0}
\renewcommand{\theequation}{A\arabic{equation}}

\section{Pseudo code} \label{sec:appendix_pseudo_code}
\begin{algorithm}[H]
\caption{Patch SVDD (train)}
  \label{algo:pseudo}
  \begin{algorithmic}[1]
  \State \textbf{Input} normal images $\left \{ \xvec \right \}$, hyperparameter $\lambda$, encoder $f_\theta$, and classifier $C_\phi$
  \For{patch $\pvec$ in $\left \{ \xvec \right \}$}   \Comment{Train the encoder}
  \ \ \ \ \State $\pvec_1 \leftarrow \texttt{RandomJitter}(\pvec)$ 
  \ \ \ \ \State $\Loss_{\text{SVDD'}} \leftarrow \left \| f_\theta (\pvec) - f_\theta (\pvec_1) \right \| _2$
  \ \ \ \ \State $\pvec_2, y \leftarrow \texttt{RandomNeighborhood}(\pvec)$  \Comment{A neighborhood in the 3 $\times$ 3 grid}
  \ \ \ \ \State $\Loss_{\text{SSL}} \leftarrow \texttt{Cross-entropy} \left (y, C_\phi \left ( f_\theta(\pvec), f_\theta(\pvec_2) \right ) \right )$
  \ \ \ \ \State Backprop $\Loss_{\text{Patch SVDD}} \leftarrow \lambda \Loss_{\text{SVDD'}} + \Loss_{\text{SSL}}$  
  \EndFor
  \State $\fbig, \fsmall \leftarrow f_\theta$   \Comment{Split encoders}
  \State $S_{\text{big}}, S_{\text{small}} \leftarrow \varnothing$   \Comment{Sets of normal features}
  \For{patch $\pvec$ in $\left \{ \xvec \right \}$}   \Comment{Patch size $K$ with stride $S$}
  \ \ \ \ \State $S_{\text{big}} \leftarrow S_{\text{big}} \cup \left \{ \fbig(\pvec) \right \}$
  \EndFor
  \For{patch $\pvec$ in $\left \{ \xvec \right \}$}   \Comment{Patch size $K$ with stride $S$}
  \ \ \ \ \State $S_{\text{small}} \leftarrow S_{\text{small}} \cup \left \{ \fsmall(\pvec) \right \}$
  \EndFor
  \State \textbf{return} $(S_{\text{big}}, S_{\text{small}})$, $(\fbig, \fsmall)$  \Comment{Normal features and trained encoders}
  \end{algorithmic}
\end{algorithm}

\begin{algorithm}[H]
\caption{Patch SVDD (test)}
  \label{algo:pseudo2}
  \begin{algorithmic}[1]
  \State \textbf{Input} query image $\xvec$, normal feature sets $(S_{\text{big}}, S_{\text{small}})$, and encoders $(f_{\text{big}}, f_{\text{small}})$
  \State Initialize $\mmap_{\text{big}}$ and $\mmap_{\text{small}}$
  \For{patch $\pvec$ in $\xvec$}     \Comment{Patch size $K$ with stride $S$}  
  \ \ \ \ \State $d \leftarrow \min_{h \in S_{\text{big}}} \left \| f_{\text{big}}(\pvec) - h \right \| _2$  \Comment{Anomaly score of a patch}
  \ \ \ \ \State Distribute $d$ to $\mmap_{\text{big}}$ of each pixel in $\pvec$
  \EndFor
  \For{patch $\pvec$ in $\xvec$}     \Comment{Patch size $K$ with stride $S$}  
  \ \ \ \ \State $d \leftarrow \min_{h \in S_{\text{small}}} \left \| f_{\text{small}}(\pvec) - h \right \| _2$  \Comment{Anomaly score of a patch}
  \ \ \ \ \State Distribute $d$ to $\mmap_{\text{small}}$ of each pixel in $\pvec$
  \EndFor
  \State $\mmap_{\text{multi}} \leftarrow \mmap_{\text{small}} \odot \mmap_{\text{big}}$ \Comment{Element-wise multiplication}
  \State $a \leftarrow \max \mmap_{\text{multi}}$ \Comment{Anomaly score}
  \State \textbf{return} $\mmap_{\text{multi}}$, $a$  \Comment{Anomaly map and anomaly score}
  \end{algorithmic}
\end{algorithm}

Algorithm~\ref{algo:pseudo} trains a hierarchical encoder using $\Loss_{\text{Patch SVDD}}$.
After the training, sets of features of normal patches are extracted using the trained multi-scale encoders.
The outputs of Algorithm~\ref{algo:pseudo} are the sets of normal features and trained encoders.
Algorithm~\ref{algo:pseudo2} performs inspection on a query image and outputs the anomaly map and anomaly score.
\newpage

\section{Results}
\subsection{Numerical results} \label{sec:appendix_numerical}
\setlength{\tabcolsep}{0.12in}
\begin{table}[H]
  \centering
  \caption{\textbf{Anomaly detection (\texttt{Det.}) and segmentation performances (\texttt{Seg.}) of proposed Patch SVDD on MVTec AD~\cite{mvtecad} dataset.} The inspection performances for each class are given in AUROC, and the average values are also reported in Table~\ref{table:anomaly_det_seg} of the main paper.}
  \vspace{0.5em}
    \begin{tabular}
    {l @{\hskip 0.1in} |c c}
    \Xhline{1pt}\\[-0.95em]
    & \multicolumn{2}{c}{\textbf{Patch SVDD}}  \\
    Classes      & \texttt{Det.} & \texttt{Seg.} \\
     
    \hline\hline
    \\[-0.9em]
    bottle         & 0.986 & 0.981 \\
    cable          & 0.903 & 0.968 \\
    capsule        & 0.767 & 0.958 \\
    carpet         & 0.929 & 0.926 \\
    grid           & 0.946 & 0.962 \\
    hazelnut       & 0.920 & 0.975 \\
    leather        & 0.909 & 0.974 \\
    metal\_nut     & 0.940 & 0.980 \\
    pill           & 0.861 & 0.951 \\
    screw          & 0.813 & 0.957 \\
    tile           & 0.978 & 0.914 \\
    toothbrush     & 1.000 & 0.981 \\
    transistor     & 0.915 & 0.970 \\
    wood           & 0.965 & 0.908 \\
    zipper         & 0.979 & 0.951 \\
    \Xhline{0.5pt}
    Average        & \textbf{0.921} & \textbf{0.957} \\
    \Xhline{0.5pt}
    \end{tabular}%
\vspace{0.5mm}
  
  \label{table:auroc_full}
\end{table}

\setlength{\tabcolsep}{0.12in}
\begin{table}[H]
  \centering
  \caption{\textbf{The effect of hierarchical encoding.} Aggregating the results from multi-scale inspection boosts the performance, and adopting hierarchical structure to the encoder is helpful as well. The plot of the data is provided in Fig.~\ref{fig:hierarchical_helps} of the main paper.}
  \vspace{0.5em}
    \begin{tabular}
    {c c @{\hskip 0.1in} |c c}
    \Xhline{1pt}\\[-0.95em]
    Hierarchical & $K$      & \texttt{Det.} & \texttt{Seg.} \\
     
    \hline\hline
    \\[-0.9em]
    \xmark & 64        & 0.810 & 0.879 \\
    \hline
    \textred{\cmark} & 64     & 0.894 & 0.932 \\
    \textred{\cmark} & 32     & 0.902 & 0.957 \\
    \textred{\cmark} & Agg. (64 \& 32)     & \textbf{0.921} & \textbf{0.957} \\
    \Xhline{0.5pt}
    \end{tabular}%
  \vspace{0.5mm}
  \label{table:auroc_hier}
\end{table}

\newpage

\subsection{Anomaly maps}
\begin{figure}[H]
    \centering
    \includegraphics[width=\linewidth]{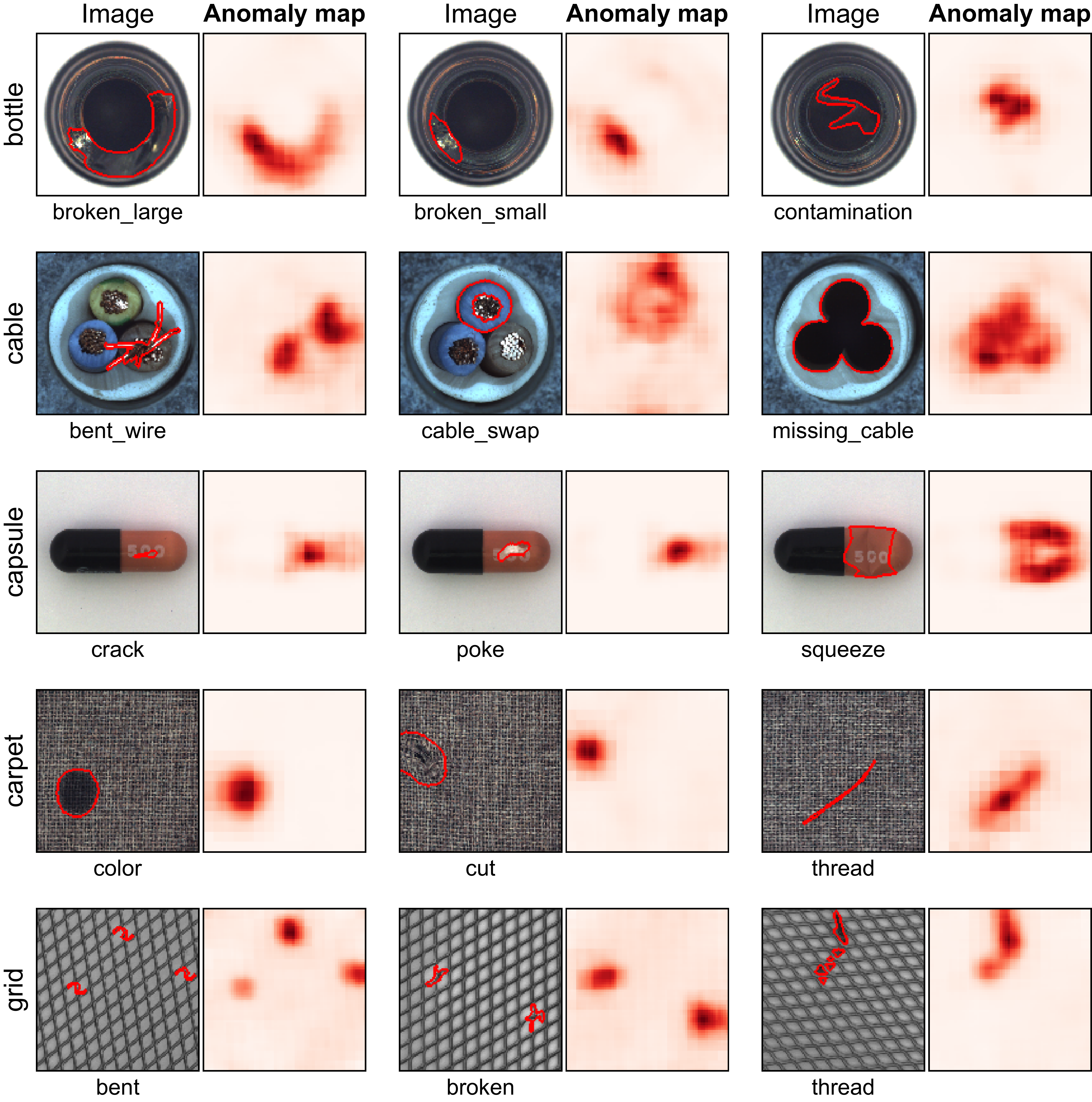}
    \vspace{-2em}
    \caption{\textbf{Anomaly maps generated by the proposed method.} Patch SVDD generates anomaly maps of the images in each class of MVTec AD~\cite{mvtecad} dataset. The ground truth defect annotations are depicted as red contours in the image, and the darker heatmap indicates higher anomaly scores. The name of the class is provided at the left of the image, and the type of the defect is indicated below the image.
    }
    \label{fig:anomaly_maps1}
\end{figure}

\clearpage
\begin{figure}[]
    \centering
    \includegraphics[width=\linewidth]{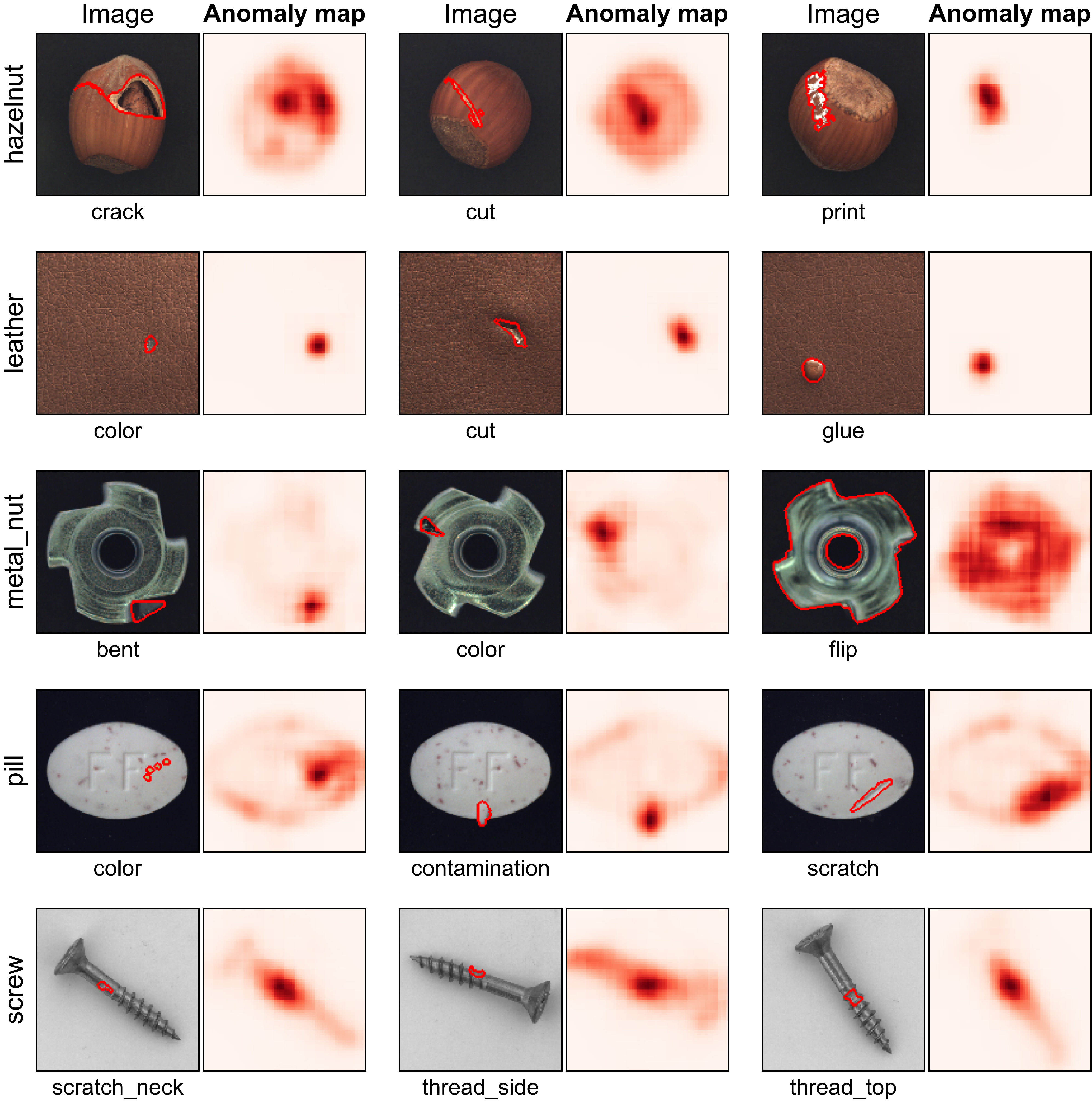}
    \vspace{-2em}
    \caption{\textbf{Anomaly maps generated by the proposed method.} Patch SVDD generates anomaly maps of the images in each class of MVTec AD~\cite{mvtecad} dataset. The ground truth defect annotations are depicted as red contours in the image, and the darker heatmap indicates higher anomaly scores. The name of the class is provided at the left of the image, and the type of the defect is indicated below the image.
    }
    \label{fig:anomaly_maps2}
\end{figure}

\clearpage
\begin{figure}[]
    \centering
    \includegraphics[width=\linewidth]{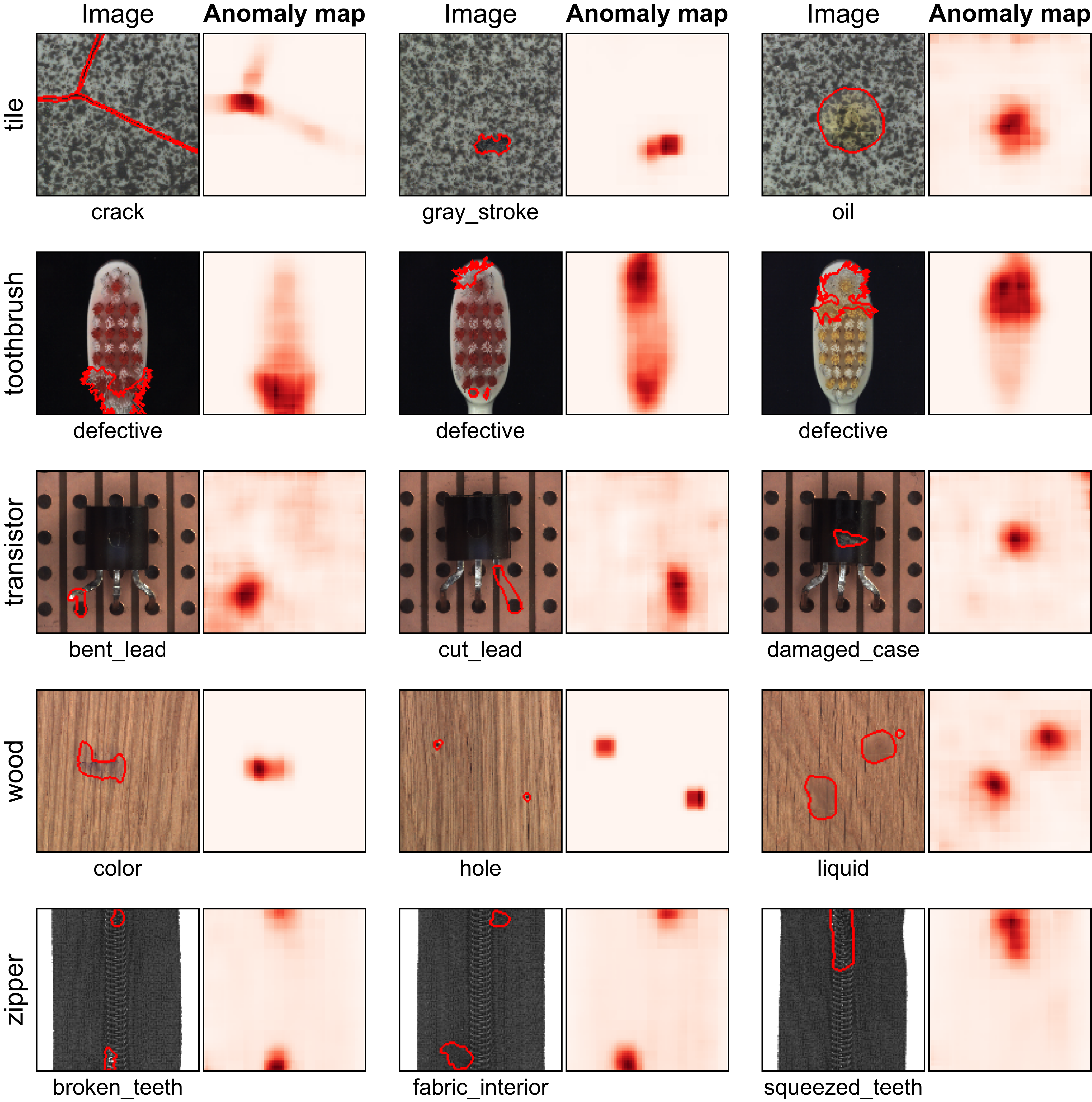}
    \vspace{-2em}
    \caption{\textbf{Anomaly maps generated by the proposed method.} Patch SVDD generates anomaly maps of the images in each class of MVTec AD~\cite{mvtecad} dataset. The ground truth defect annotations are depicted as red contours in the image, and the darker heatmap indicates higher anomaly scores. The name of the class is provided at the left of the image, and the type of the defect is indicated below the image.
    }
    \label{fig:anomaly_maps3}
\end{figure}

\clearpage

\section{Implementation details} \label{sec:appendix_implementation}
\subsection{Dataset}
The dataset in the study, MVTec AD~\cite{mvtecad}, consists of 15-class industrial images.
Each class is categorized as either an \textit{object}\footnote{bottle, cable, capsule, hazelnut, metal\_nut, pill, screw, toothbrush, transistor, and zipper} or \textit{texture}\footnote{carpet, grid, leather, tile, and wood}.
Each class contains 60 to 390 normal train images and 40 to 167 test images.
Test images include both normal and abnormal examples, and the defects of the abnormal images are annotated at the pixel level in the form of binary masks.
We downsampled every image to a resolution of 256 $\times$ 256.
Gray-scale images are converted to RGB images by replicating the single channel to three.
No data augmentation method (e.g., horizontal flip, rotation) was used for the training.

\subsection{Networks}
Two neural networks are used throughout the study: an encoder and a classifier.
The encoder is composed of convolutional layers only.
The classifier is a two-layered MLP model having 128 hidden units per layer, and the input to the classifier is a subtraction of the features of the two patches.
The activation function for both networks is a LeakyReLU~\cite{relu} with a $\alpha=0.1$.
Please refer to the code\footnote{\url{https://github.com/nuclearboy95/Anomaly-Detection-PatchSVDD-PyTorch}} for the detailed architecture of the networks.

As proposed in Section 3.3 of the main paper, the encoder has a hierarchical structure.
The receptive field of the encoder is $K=64$, and that of the embedded smaller encoder is $K=32$.
Patch SVDD divides the images into patches with a size $K$ and a stride $S$.
The values for the strides are $S=16$ and $S=4$ for the encoders with $K=64$ and $K=32$, respectively.

\subsection{Environments}

The experiments throughout the study were conducted on a machine equipped with an Intel i7-5930K CPU and an NVIDIA GeForce RTX 2080 Ti GPU.
The code is implemented in python 3.7 and PyTorch~\cite{pytorch}.

\end{document}